\documentclass[10pt, a4paper]{article}

\usepackage[]{lrec-coling2024} 

\name{Weicheng Ren$^{1,2}$, Zixuan Li$^{2}$\sthanks{\ \ Corresponding authors}, Xiaolong Jin$^{1,2}$$^{*}$, Long Bai$^{2}$, Miao Su$^{1,2}$,  \\ \large{\textbf{Yantao Liu$^{1,2}$, Saiping Guan$^{2}$, Jiafeng Guo$^{1,2}$, Xueqi Cheng$^{1,2}$}}}

\address{$^{1}$School of Computer Science and Technology, University of Chinese Academy of Sciences;\\ $^{2}$Key Lab of Network Data Science and Technology,\\ Institute of Computing Technology, Chinese Academy of Sciences. \\
         \{renweicheng21b, lizixuan, jinxiaolong, bailong18b, sumiao22z\}@ict.ac.cn \\ 
         \{liuyantao22s, guansaiping, guojiafeng, cxq\}@ict.ac.cn\\}

\title{Nested Event Extraction upon Pivot Element Recognition}

\abstract{
Nested Event Extraction (NEE) aims to extract complex event structures where an event contains other events as its arguments recursively. Nested events involve a kind of Pivot Elements (PEs) that simultaneously act as arguments of outer-nest events and as triggers of inner-nest events, and thus connect them into nested structures. This special characteristic of PEs brings challenges to existing NEE methods, as they cannot well cope with the dual identities of PEs. Therefore, this paper proposes a new model, called PerNee, which extracts nested events mainly based on recognizing PEs. Specifically, PerNee first recognizes the triggers of both inner-nest and outer-nest events and further recognizes the PEs via classifying the relation type between trigger pairs. The model uses prompt learning to incorporate information from both event types and argument roles for better trigger and argument representations to improve NEE performance. Since existing NEE datasets (e.g., Genia11) are limited to specific domains and contain a narrow range of event types with nested structures, we systematically categorize nested events in the generic domain and construct a new NEE dataset, called ACE2005-Nest. Experimental results demonstrate that PerNee consistently achieves state-of-the-art performance on ACE2005-Nest, Genia11, and Genia13. The ACE2005-Nest dataset and the code of the PerNee model are available at \url{https://github.com/waysonren/PerNee}.
 \\ \newline  }

\pgfplotsset{compat=1.18}

\begin{document}

\maketitleabstract


\section{Introduction}

\begin{figure}
    \centering
    \includegraphics[width=1\linewidth]{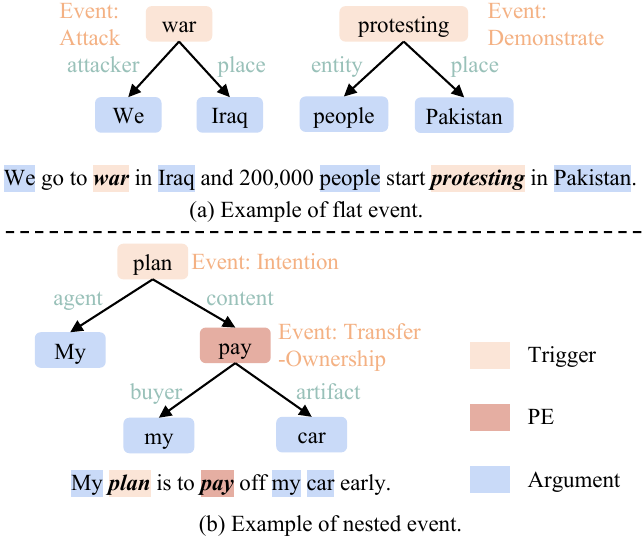}
    \setlength{\abovecaptionskip}{0pt}
    \caption{Examples of flat (a) and nested (b) events.}
    \setlength{\belowcaptionskip}{0pt}
    \label{example}
\end{figure}

Event Extraction (EE), as an important task in information extraction, aims to extract event triggers and their corresponding arguments from sentences. Traditional EE implicitly assumes that all events in the same sentence have flat structure, thus called Flat Event Extraction (FEE). However, there also exists a kind of nested structures where an event contains other events as its arguments recursively. Therefore, Nested Event Extraction (NEE) as a new information extraction task has recently attracted attention~\cite{trieu2020deepeventmine,cao2022oneee}. Figure~\ref{example} (a) and (b) illustrate two examples of both flat and nested events, correspondingly. NEE holds immense importance in attaining a profound semantic understanding and acquiring a comprehensive perspective of the event structure.

In the case of nested events, events are connected via a kind of special elements that simultaneously act as arguments of outer-nest events and as triggers of inner-nest events. This kind of elements play as pivots in the nested event structures, thus called Pivot Elements (PEs) in this paper. As shown in Figure~\ref{example}(b), ``pay'' is a PE, which serves as the trigger of the inner-nest event \texttt{Transfer-Ownership} and as an argument of the outer-nest event \texttt{Intention}. Through the PE ``pay'', these two events are connected to form a nested event structure. Therefore, the key for the NEE task is to recognize this kind of PEs.

However, the dual identities of PEs present challenges to existing NEE methods~\cite{lin2020joint, cao2022oneee}. These methods typically employ two separate modules to extract triggers and arguments, and recognize those overlapping ones as PEs. However, due to their more trigger-like characteristics of PEs, it is difficult for the argument extraction module to recognize them as the arguments of outer-nest events, which affects the performance of those existing methods on NEE.

To address this challenge, we propose PerNee, a novel model for the NEE task via better recognizing PEs. Unlike existing methods, PerNee transfers the identification problem of the argument identities of PEs to a classification problem of relations between trigger pairs within the same sentences. Specifically, PerNee utilizes the label names of event types and argument roles as prompts, which are prepended to the sentences. It then employs a BERT-based network to encode the sentences along with these prompts, generating contextual representations enriched with the information of event types and argument roles. Next, PerNee recognizes triggers and regular arguments (i.e., entities that are definitely not PEs) by employing two separate Feedforward Neural Networks (FNNs) combined with a Conditional Random Field (CRF) layer. Finally, PerNee identifies every regular argument corresponding to its trigger and further determines its role by generating and classifying pairs between triggers and regular arguments using an FNN. Simultaneously, by generating and classifying the pairs of triggers based on another FNN, it recognizes from the set of triggers, if any, every PE as well as the trigger of its corresponding outer-nest event and its role therein. By so doing, the nested event structure contained in the input sentence is identified.

There are several event extraction datasets containing nested events (e.g., Genia11~\citelanguageresource{kim2011overview}, Genia13~\citelanguageresource{kim2013genia}). However, these existing datasets primarily focus on the medical domain and have a narrow range of event types that can introduce nested structures. For instance, in Genia11, only some of the \texttt{Regulation} events exhibit nested structures. In contrast, the generic domain contains a diverse array of event types that can introduce nested events, such as \texttt{Intention}, \texttt{Belief}, and \texttt{Statement}. To address these limitations, we systematically categorize nested events in the generic domain into different types and create a new NEE dataset, ACE2005-Nest, based on the widely used benchmark dataset ACE2005 for FEE. ACE2005-Nest contains 14 event types that can introduce nested structures in the generic domain.

Our contributions can be summarized as follows:

\begin{itemize}
    \item We propose PerNee for the NEE task, which extracts nested events mainly based on recognizing PEs. By classifying the relations between trigger pairs, PerNee significantly enhances the accuracy of PE extraction.
    \item We systematically categorize nested events in the generic domain and construct a new NEE dataset, ACE2005-Nest, which can serve as a valuable resource to advance the NEE task in the generic domain.
    \item Experimental results demonstrate that the PerNee model consistently outperforms existing baselines on ACE2005-Nest, Genia11, and Genia13, demonstrating its effectiveness in both FEE and NEE tasks.

\end{itemize}

\section{Related Work}

\subsection{Nested Event Extraction}

Some existing studies tackle NEE using methods actually for overlapping events~\cite{yang2019exploring,li2020event, sheng2021casee, cao2022oneee}, as NEE can be seen as a specific type of overlapping events, where triggers and arguments overlap. For example, \citet{cao2022oneee} proposed OneEE to address both overlapping and nested events. PEs are recognized in both the trigger recognition module and the argument recognition module, which handles the overlapping issue between triggers and arguments, thereby addressing NEE. In a similar manner, some existing FEE methods~\cite{nguyen2019one, raffel2020exploring, wadden2019entity, lin2020joint, lu2022unified, shi-etal-2023-hybrid} can be adapted to address NEE by treating PEs as both triggers and regular arguments and recognizing the overlapping ones as PEs. 

However, these methods face difficulties in coping with the dual identities of PEs. They simply treat PEs as regular arguments and extract them within the argument extraction module, neglecting their trigger-like characteristics, which brings challenges to argument extraction.

\subsection{NEE Datasets}

In the medical domain, there are several NEE datasets available. Genia11~\citelanguageresource{kim2011overview} is a medical domain event extraction dataset, containing a total of 9 event types. Among these event types, \texttt{Regulation}, \texttt{Positive Regulation}, and \texttt{Negative Regulation} are 3 event types that can involve other events as arguments. Based on Genia11, Genia13~\citelanguageresource{kim2013genia} introduces additional event types such as \texttt{Phosphorylation} that can introduce nested events. Besides, in the Cancer Genetics dataset~\cite{pyysalo2013overview} and Pathway Curation dataset~\cite{ohta2013overview}, the \texttt{Regulation} event type is prominent for introducing nested event structures.

Above all, in existing NEE datasets, nested events are mainly concentrated in limited event types like \texttt{Regulation}, with a predominant focus on the medical domain. However, in the generic domain, nested events are widespread with a diverse range of types, indicating a need for generic domain NEE datasets.

\section{Problem Formulation}

Given a sentence $X$, the NEE task aims to extract the events therein, including their triggers and arguments, and further identify the specific roles of all extracted arguments and, if any, the nested structures between events. Let $E = \{e_1, e_2, ..., e_k\}$ be the set of events contained in $X$. Each event $e_i$ ($1\leq i\leq k$) is represented as a 4-tuple $(\tau_i, t_i, A_i, R_i)$, where $\tau_i$ is its type and $t_i$ is its trigger associated with $\tau_i$, indicating its occurrence; $A_i$ and $R_i$ are the sets of its arguments and their corresponding roles, respectively. For each $e_i$, the $l^{th}$ argument $a_{i}^{l}\in A_i $ is associated with a corresponding role in $R_i$.

The nested event structures, if any, in $E$ that can essentially be characterized by a PE set $P = \{t_i|\exists{j}, t_i \in A_j\}$. In this view, the NEE task involves the following subtasks:

\textbf{Trigger Recognition}: Given the sentence $X$, it is to recognize all triggers $T = \{t_1, t_2, ..., t_k\}$ therein and further determine their respective event types.

\textbf{Regular Argument Extraction}: Given the sentence $X$ and a trigger $t_i \in T$, this subtask is to extract the set of its arguments excluding PEs, $A_i =\{a_{i}^{1}, a_{i}^{2}, ..., a_{i}^{l}, ...\}$ and further determine their respective roles in $R_i$.

\textbf{Pivot Element Recognition}: Given the sentence $X$ and a trigger $t_i \in T$, the goal is to identify whether or not there exists another trigger $t_j\in T$, $t_i$ is one of its argument and, if so, further determine its role.

\section{The PerNee Model}

\begin{figure*}
    \centering
    \includegraphics[width=1\linewidth]{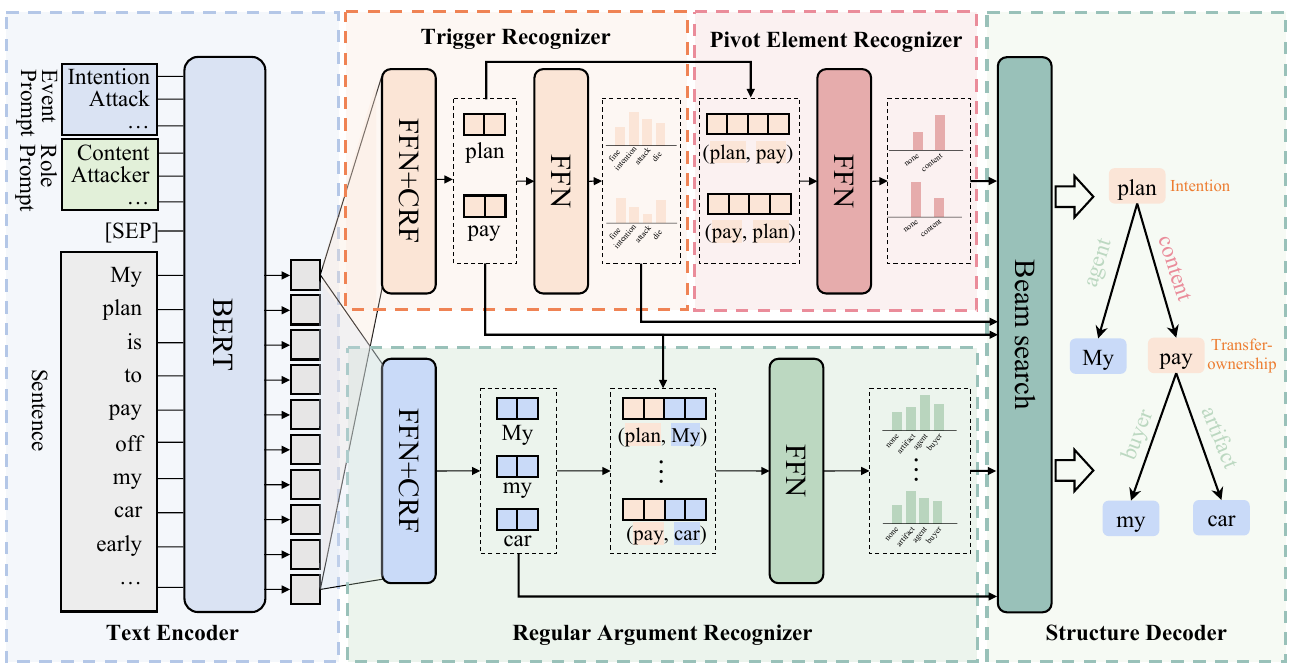}
    \setlength{\abovecaptionskip}{0.cm}
    \caption{The overall framework of the PerNee model.}
    \setlength{\belowcaptionskip}{0pt}
    \label{fig:model}
\end{figure*}

In this section, we will introduce the framework of PerNee. As shown in Figure~\ref{fig:model}, it mainly contains five modules. The text encoder encodes the sentence with prompts to obtain the representations of all words therein. Based on these representations, the trigger recognizer and the regular argument recognizer recognize triggers and regular arguments, respectively. Next, the pivot element recognizer is adopted to recognize, if any, all PEs. Based on the extracted elements, the structure decoder explores possible event structures using beam search to generate events with the highest global score.

\subsection{The Text Encoder}
This module aims to obtain the representation for each word within a given sentence $X$. In order to acquire the word representation enriched with a contextual understanding of the event types and argument roles, we prepend the label names of all event types and argument roles as prompts to $X$. This, in turn, enhances the model's perception of event schema. Some related papers~\cite{lu2022unified, wang-etal-2022-query, Lou_Lu_Dai_Jia_Lin_Han_Sun_Wu_2023} have demonstrated that introducing label information of event types and argument roles can improve the ability of the model to perceive the information to be extracted.

Following \citet{brown2020language, schick2020exploiting}, we use [EVENT] and [ROLE] as placeholder separators (abbreviated as [$\mathcal{T}$] and [R] hereafter) to concatenate the label names of event types $\tau_{i}$ and argument roles $r_{i}$. Finally, the input of the text encoder is:
\begin{equation*}
    [\mathcal{T}] \tau_{1} [\mathcal{T}] \tau_{2} ...[\mathcal{T}] \tau_{n} [R] r_{1} [R] r_{2} ... [R] r_{n} [SEP] X
\end{equation*}

Next, the input is encoded through a pre-trained BERT model~\cite{devlin2018bert}.
As BERT tokenizes each word into several subword pieces (e.g., ``blowdryers'' → ``blow'', ``\#\#dr'', ``\#\#yers''), we obtain its representation by computing the average of the representations of those corresponding subword pieces.

Finally, this module generates the representations of all words in $X$, denoted as $\mathbf{H}=\{\mathbf{h_1}, \mathbf{h_2}, ..., \mathbf{h_n}\}$.

\subsection{The Trigger Recognizer}
\label{trigger_recognizer}

This module aims to recognize triggers, which contains two steps: an identification step to identify triggers and a classification step to obtain, for each trigger, the label scores of corresponding event types. 

The trigger identification can be formulated as a sequence labeling problem. Specifically, the module takes word representations in each sentence as its input and calculates a score vector for each word using an FNN. Each value in the vector represents the score of a specific tag corresponding to the BIO tag schema. To capture the dependencies among predicted tags, a CRF layer is utilized to ensure the validity of certain tag sequences. For instance, an \textit{I-Intention} tag should not follow a \textit{B-Attack} tag. The trigger tag sequence corresponding to the sentence is obtained as $\mathbf{\hat{z}^t}$. Inspired by~\citet{lample2016neural}, the objective is to maximize the log-likelihood of the gold-standard tag sequence. Thus, the loss of trigger identification is defined as:
\begin{equation}
\label{eq:ti}
\mathcal{L}^{t}_{1} = \log \sum_{\mathbf{\hat{z}}^{t} \in Z^{t}}{e^{s(\mathbf{H},\mathbf{\hat{z}}^{t})}} - s(\mathbf{H},\mathbf{z}^{t}),
\end{equation}
where $s$ denotes the tag sequence scoring function, $\mathbf{z}^{t}$ represents the golden trigger tag sequence, and $Z^{t}$ represents the set of all possible trigger tag sequences for a given sentence.

In the classification step, since the identified triggers may contain several words, the representation of the $i^{th}$ identified trigger is obtained by averaging its word representations, denoted as $\mathbf{t_{i}}$. To obtain its corresponding event type, another FNN is employed to calculate type label scores as $\mathbf{\hat{y}}^{t}_{i} = FNN(\mathbf{t_i})$. 

For trigger classification, the objective is to minimize the following cross-entropy loss:

\begin{equation}
\mathcal{L}^{t}_{2} = - \frac{1}{N^{t}} \sum_{i=1}^{N^{t}} \mathbf{y}^{t}_{i} \log \mathbf{\hat{y}}^{t}_{i},    
\end{equation}
where $N^{t}$ and $\mathbf{y}^{t}_{i}$ represent the number of triggers and the true label vector, respectively. Therefore, the training loss of the trigger recognizer is defined as:
\begin{equation}
\mathcal{L}^{t} = \mathcal{L}^{t}_{1} + \mathcal{L}^{t}_{2}.
\end{equation}

\subsection{The Regular Argument Recognizer}

Considering the trigger-like characteristics of PEs and their notable differences from regular arguments (i.e., entities), jointly recognizing PEs and regular arguments may affect the performance of argument recognition. Therefore, this regular argument recognizer focuses only on extracting regular arguments. It involves two steps: an identification step to extract regular arguments and a classification step to obtain the label scores of role types. 

In the identification step, this module employs an FNN followed by a CRF layer to generate tag sequences for regular arguments. Similar to Equation~\ref{eq:ti}, the loss of regular argument identification is defined as:
\begin{equation}
\mathcal{L}^{a}_{1} = \log \sum_{\mathbf{\hat{z}}^{a} \in Z^{a}}{e^{s(\mathbf{H},\mathbf{\hat{z}}^{a})}} - s(\mathbf{H},\mathbf{z}^{a}),    
\end{equation}
where $\mathbf{\hat{z}^a}$, $\mathbf{z}^{a}$, and $Z^{a}$  represent the predicted regular argument tag sequence, the golden regular argument tag sequence, and the set of all possible regular argument tag sequences for a given sentence, respectively.

In the classification step, role types of arguments are determined by establishing relations between triggers and regular arguments. Given a trigger and a regular argument, the representation of the trigger-argument pair is calculated by concatenating the representations of the identified trigger and regular argument, denoted as $[\mathbf{t_i};\mathbf{a_j}]$. Then, another FNN is employed to calculate the score vector of the trigger-argument pair, denoted as $\mathbf{\hat{y}}^{a}_{i,j} = FNN([\mathbf{t_i};\mathbf{a_j}])$, which represents the role type scores for the identified regular argument.

For trigger-argument pair classification, the objective is to minimize the following cross-entropy loss:
\begin{equation}
\mathcal{L}^{a}_{2} = - \frac{1}{N^{a}} \sum_{i=1}^{N^{a}} \mathbf{y}^{a}_{i,j} \log \mathbf{\hat{y}}^{a}_{i,j},    
\end{equation}
where $N^{a}$ and $\mathbf{y}^{a}_{i,j}$ represent the number of trigger-argument pairs and the true label vector, respectively. Therefore, the training loss of the regular argument recognizer is defined as:

\begin{equation}
\mathcal{L}^{a} = \mathcal{L}^{a}_{1} + \mathcal{L}^{a}_{2}.    
\end{equation}

\subsection{The Pivot Element Recognizer}
\label{sec:per}
The nested events arise when one event serves as an argument of another event. To recognize nested events, it is crucial to recognize PEs. However, PEs bring challenges to existing methods due to their dual identities. Considering the trigger-like characteristics of PEs, PerNee first identifies the trigger identities of PEs via the trigger recognizer as mentioned in Section~\ref{trigger_recognizer}. Then, in the pivot element recognizer, PerNee further identifies the argument identities of PEs by transferring the identification problem to a classification problem of the relations between trigger pairs within the same sentence. By doing so, recognizing PEs can be transferred to discovering the argument relations between the trigger pairs, hereby helping the model avoid confusion arising from the dual identities of PEs.

Specifically, given the set $T$ of all the extracted triggers in a sentence, PerNee first generates the candidate trigger pairs $\{(t_i, t_j)| t_i, t_j \in T\}$. Note that if the trigger pair $(t_i, t_j)$ is added to candidate trigger pairs, the trigger pair $(t_j, t_i)$ is also included. Then, the representations of the triggers are concatenated to form the representation of the trigger pair, represented as $[\mathbf{t_i};\mathbf{t_j}]$. An FNN is employed to calculate the score vector of the trigger pair, denoted as $\mathbf{\hat{y}}^{p}_{i,j} = FNN([\mathbf{t_i};\mathbf{t_j}])$, which represents the role type scores for the PE in the corresponding outer-nest event.

For trigger-trigger pair classification, the objective is to minimize the following cross-entropy loss: 
\begin{equation}
\mathcal{L}^{p} = - \frac{1}{N^{p}} \sum_{i=1}^{N^{p}} \mathbf{y}^{p}_{i,j} \log \mathbf{\hat{y}}^{p}_{i,j},    
\end{equation}
where $N^{p}$ and $\mathbf{y}^{p}_{i,j}$ represent the number of trigger-trigger pairs and the true label vector, respectively.

Finally, the joint objective function during training is optimized by minimizing the following loss function:
\begin{equation}
\mathcal{L} = \mathcal{L}^{t} + \mathcal{L}^{a} + \mathcal{L}^{p}.    
\end{equation}

\subsection{The Structure Decoder}

In the prediction stage, we first extract the elements and their corresponding score vectors based on the above modules and subsequently employ a beam search-based strategy to decode the globally optimal event structure, following~\cite{lin2020joint}. This approach aims to achieve global best extraction results instead of local ones. In this context, event structures are represented as graphs in which triggers and regular arguments serve as nodes, connected by edges denoting their relations. The score for a given graph $g$ is computed as:
\begin{equation}
score(g) = \sum_{i=0}^{N^{v}}{s(v_{i})}+\sum_{i=0}^{N^{\ell}}{s(\ell_{i})},    
\end{equation}
where $s(v_{i})$, $s(\ell_{i})$ represent the scores of node types and edge types, and $N^{v}$, $N^{\ell}$ denote the number of nodes and edges. Note that all scores are normalized within the nodes or edges.

Beam search is used to iteratively extend nodes and edges with a beam set of size $\theta$. The extension process involves selecting the top $k$ most likely labels for both nodes and edges. After extending nodes and edges, a set of candidate graphs is obtained, denoted as $G =\{g_1, g_2, ..., g_n\}$. The graph with the highest score is then selected from this set:
\begin{equation}
g_{best} = \mathop{\arg\max}\limits_{g_k \in G} (score(g_k)), k = 1, 2, ..., n.    
\end{equation}

\section{The ACE2005-Nest Dataset}

To address the limitations of existing NEE datasets, such as Genia11, which are domain-specific and have a limited range of event types that can introduce nested structures, we construct a new NEE dataset in the generic domain, building upon the ACE2005 dataset\footnote{https://catalog.ldc.upenn.edu/LDC2006T06} (a widely used source for FEE). It contains 8 event categories, 33 sub-categories, and 35 argument roles, derived from news, broadcasts, and conversations. Based on ACE2005, we discover extra event types that can introduce nested structures and their associated argument roles. We then annotate instances of these new event types based on the original events.

\subsection{Nested Event Schema Discovery}

Building upon the existing event annotations in the ACE2005 dataset, we discover the nested event schema as follows: First, triggers that may cause nested structures are identified; Then, the inner-nest events (i.e., PEs) are identified as well as their relevant arguments, such as agent and time. After that, we build up the connections between the triggers and their respective arguments. 

To categorize event types and determine the frame semantics descriptions, some established resources are referred to, including WordNet~\cite{fellbaum2010wordnet}, FrameNet~\cite{baker1998berkeley}, and FactBank~\cite{sauri2009factbank}. These resources provide valuable insights into verb classification and frame semantics. With this knowledge, we systematically define various types of triggers that have the potential to introduce nested structures. Based on our analysis, these triggers can be classified into 7 categories and 14 sub-categories, as shown in Table~\ref{tab:schema}.

\begin{table}[ht]
\small
\centering
\begin{tabular}{ccc}
\toprule
\textbf{Event Types} & \textbf{Subtypes}  & \textbf{Trigger Examples} \\ 
\midrule
\multirow{2}{*}{\tabincell{c}{Statement}}
 & Oral &  say, speak \\ \cmidrule{2-3}
 & Written  & write, report\\ 
 \midrule
\multirow{3}{*}{\tabincell{c}{Idea}} 
 & \multirow{1}{*}{Belief}  & believe, think \\ \cmidrule{2-3}
 & Attitude  & oppose, agree \\  \cmidrule{2-3}
 & \multirow{1}{*}{Doubt}  & wonder, doubt \\ 
 \midrule
\multirow{3}{*}{Knowledge}
 & \multirow{1}{*}{Aware} & know, aware \\ \cmidrule{2-3}
 & \multirow{1}{*}{Perception}  & see, hear \\ \cmidrule{2-3}
 & \multirow{1}{*}{Inference}  & mean, indicate \\ 
 \midrule
\multirow{2}{*}{Sentiment}
 & \multirow{1}{*}{Preference}  & like, hate \\ \cmidrule{2-3}
 & \multirow{1}{*}{Emotion}  & worry, fear \\ 
 \midrule
\multirow{2}{*}{Instruction}
 & \multirow{1}{*}{Command}  & order, instruct \\ \cmidrule{2-3}
 & \multirow{1}{*}{Demand}  & require, ask \\ 
 \midrule
\multirow{1}{*}{Judgement} & - & accuse, blame \\ 
\midrule
\multirow{1}{*}{Intention} & -  & plan, want \\ 
\bottomrule
\end{tabular}
\caption{Event types in the generic domain that can introduce nested events and their corresponding trigger examples.}
\label{tab:schema}
\end{table}

\subsection{Data Analysis}

ACE2005-Nest is divided into the train, dev, and test sets following pre-processing of \citet{wadden2019entity}. We conduct an analysis of ACE2005-Nest, along with the other two NEE datasets, Genia11 and Genia13, as shown in Table~\ref{tab:dataset}. It reveals that in ACE2005-Nest, approximately 25\% of the sentences with events contain nested events, while in Genia11 and Genia13, the account is 39\% and 49\%. Besides, ACE2005-Nest significantly surpasses Genia11 and Genia13 in terms of the number of event types capable of introducing nested events. While Genia11 and Genia13 only have 3 and 5 such event types, ACE2005-Nest has 14, indicating that ACE2005-Nest exhibits greater diversity in event types capable of introducing nested events.

\begin{table}[t]
\centering
\small
\resizebox{\linewidth}{!}{
\begin{tabular}{p{1.1cm} lcccccc}
\toprule
                            &         & \#S.     &\#S.E.  & \#S.N.E. & \#E.T. & \#E.T.N. \\ \midrule
\multirow{3}{1.1cm}{ACE2005-Nest}      
                            & Train   & 19,204         & 3,342     & 778 & \multirow{3}{*}{47} & \multirow{3}{*}{14}  \\
                            & Dev     & 901          & 327     & 103 & &  \\
                            & Test    & 676          & 293      & 112 & &  \\ \midrule

\multirow{3}{*}{Genia11} & Train     & 8,722        & 3,707     & 1,464 & \multirow{3}{*}{9} & \multirow{3}{*}{3}    \\
                          & Dev    & 1,090       & 474     & 167 & &     \\
                         & Test   & 1,091         & 456     & 173 & &     \\ \midrule
\multirow{3}{*}{Genia13} & Train        & 4,000     &1,574   & 795 & \multirow{3}{*}{13} & \multirow{3}{*}{5}     \\
                            & Dev      & 500         &189    & 90 & &   \\
                            & Test      & 500        &201    & 85 & &    \\ 
                        \bottomrule
\end{tabular}
}
\caption{Statistics of the datasets. ``\#S.'', ``\#S.E.'', ``\#S.N.E.'', ``\#E.T.'', and ``\#E.T.N.'' denote the numbers of sentences, sentences with events, sentences with nested events, event types, and event types that can introduce nested events, respectively.}
\label{tab:dataset}
\end{table}

\begin{figure}[ht]
    \centering
    \includegraphics[width=1\linewidth]{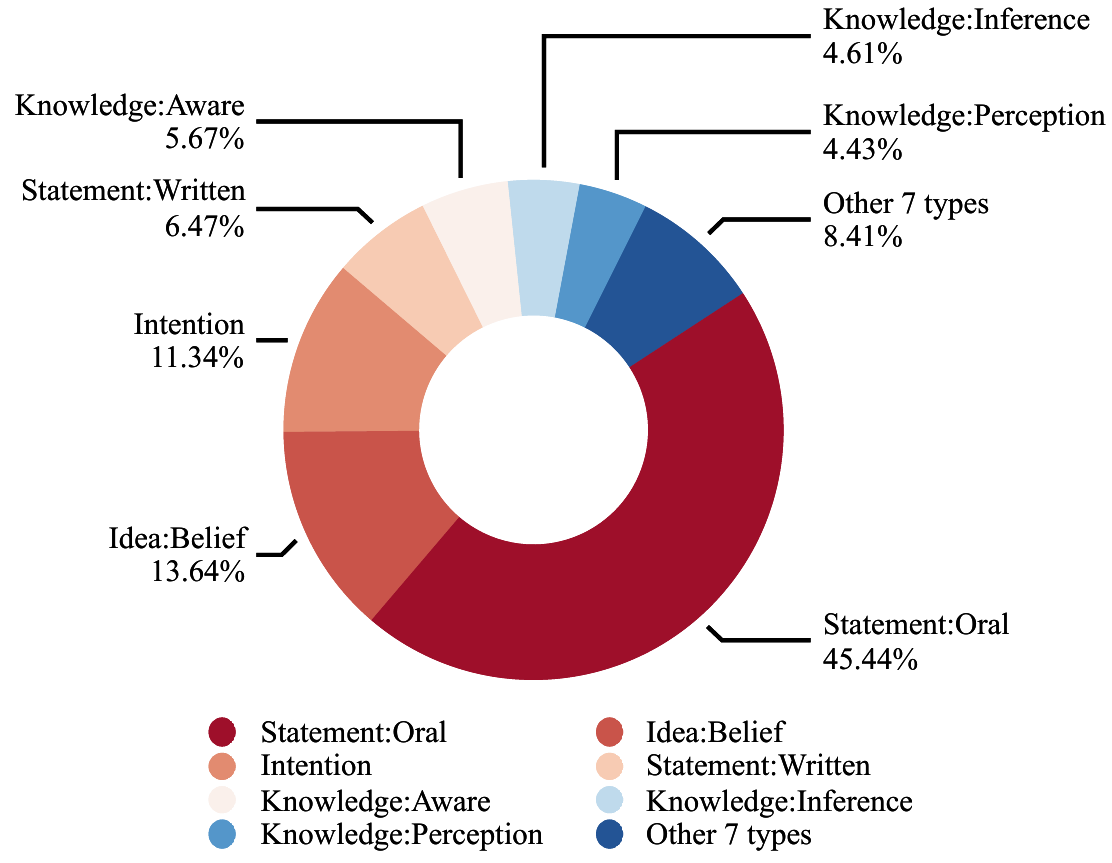}
    \setlength{\abovecaptionskip}{0pt}
    \caption{Analysis of the proportions of event types capable of introducing nested events.}
    \setlength{\belowcaptionskip}{0pt}
    \label{fig:stat}
\end{figure}

Additionally, we conduct a detailed analysis of the proportions of event types that may introduce nested events, as shown in Figure~\ref{fig:stat}. The results show that \texttt{Statement:Oral}, \texttt{Idea:Belief} and \texttt{Intention} are the top three event types that may introduce nested structures with the highest number of occurrences, accounting for 45.54\%, 13.64\%, and 13.64\%, respectively.

Besides, the ACE2005-Nest dataset also has some shortcomings: (1) The coverage breadth of event types capable of introducing nested events is insufficient. Nested events are a common phenomenon in natural language, and our current classification is based on statistical analysis during the annotation process and referencing some resources such as WordNet~\cite{fellbaum2010wordnet}, FrameNet~\cite{baker1998berkeley}, and FactBank~\cite{sauri2009factbank}. However, this is still a preliminary exploration, and the relevant definitions need further refinement and supplementation. (2) ACE2005-Nest is annotated based on ACE2005. Due to inherent noise in ACE2005 and variations in the standards among annotators during the labeling process, additional noise may be introduced.

\section{Experiments}

\subsection{Experimental Setup}

{\bf Datasets.} We conduct experiments on ACE2005-Nest, Genia11 and Genia13. The statistics of datasets are shown in Table \ref{tab:dataset}. Since the golden annotations for the test set of Genia11\footnote{\scriptsize https://2011.bionlp-st.org/home/genia-event-extraction-genia} and Genia13\footnote{ \scriptsize https://bionlp.dbcls.jp/projects/bionlp-st-ge-2013/wiki/Overview} are not released, we follow \citet{cao2022oneee} to split the train and dev sets into train/dev/test sets with a proportion of 8:1:1 randomly.

{\bf Evaluation Metrics.} For evaluation, we use the same criteria of previous work~\cite{cao2022oneee} and two new criteria. The traditional criteria are as follows: 1) Trigger Identification (\textbf{TI}): A trigger is correctly identified when its span matches the golden label; 2) Trigger Classification (\textbf{TC}): A trigger is correctly classified when it is identified correctly and its event type matches the golden label; 3) Argument Identification (\textbf{AI}): An argument is correctly identified when its event type matches the golden label and its predicted span matches the golden label; 4) Argument Classification (\textbf{AC}): An argument is correctly classified when it is identified correctly, and its role type matches the golden label. Besides, to evaluate the performance of PE recognition, two new criteria are added: 5) Pivot Element Identification (\textbf{PEI}): A PE is correctly identified when its outer-nest event type is recognized correctly, and its predicted span matches the golden label; 6) Pivot Element Classification (\textbf{PEC}): A PE is correctly classified when it is identified correctly, and its role in outer-nest events matches the golden label. We report the average F-measure (F1) scores across five runs for each of these metrics.

{\bf Baselines.}
For evaluation, we choose OneIE~\cite{lin2020joint}, UIE~\cite{lu2022unified}, PLMEE~\cite{yang2019exploring}, CasEE~\cite{sheng2021casee}, HDGSE~\cite{shi-etal-2023-hybrid}, and OneEE~\cite{cao2022oneee} as the baseline models. OneIE, UIE, and HDGSE were initially designed for FEE and have exhibited robust performance. We adapt these models to NEE by considering spans that are simultaneously identified as triggers and regular arguments as PEs. Furthermore, aligning with previous work~\cite{cao2022oneee}, PLMEE, CasEE, and OneEE are selected as our baselines that have demonstrated strong performance in overlapping and nested event extraction tasks.

\begin{table*}[ht]
\centering
\resizebox{\linewidth}{!}{
\begin{tabular}{lcccccccccccc} 
 \toprule
   & \multicolumn{4}{c}{ACE2005-Nest} & \multicolumn{4}{c}{Genia11} & \multicolumn{4}{c}{Genia13}\\ \cmidrule(r){2-5} \cmidrule(r){6-9} \cmidrule(r){10-13}
   & TI & TC & AI & AC &  TI & TC & AI & AC & TI & TC & AI & AC  \\ 
  \midrule
PLMEE~\cite{yang2019exploring} & 60.8 & 58.5 & 45.2 & 44.3 & 65.4 & 63.5 & 57.1 & 56.8 & 72.6 & 69.9 & 65.7 & 64.9 \\
CasEE~\cite{sheng2021casee} & 71.9 & 67.2 & 47.8 & 46.3 & 70.8 & 68.2 & 60.2 & 59.8 & 78.9 & 75.1 & 67.8 & 67.1 \\
UIE~\cite{lu2022unified} & 70.0 & 66.0 & 50.7 & 48.2 & 70.4 & 68.5 & 60.7 & 59.2 & 75.5 & 72.2 & 66.7 & 66.3 \\
HDGSE~\cite{shi-etal-2023-hybrid} & 72.3 & 69.5 & 49.9 & 46.9 & 71.5 & 69.2 & 61.3 & 59.9 & 77.1 & 76.2 & 69.4 & 67.7 \\
OneIE~\cite{lin2020joint} & 70.9 & 68.3 & 53.1 & 51.3 & 71.0 & 68.8 & 58.7 & 57.4 & 78.7 & 75.2 & 66.4 & 64.6 \\
OneEE~\cite{cao2022oneee} & 72.7 & 69.9 & 51.3 & 48.4 & 71.3 & 69.2 & 61.9 & 60.8 & 78.6 & 76.4 & 70.8 & 68.1 \\
\cdashlinelr{1-13}

PerNee & \textbf{72.8} & \textbf{70.0} & \textbf{55.5} & \textbf{53.8} & \textbf{71.8} & \textbf{69.5} & \textbf{62.3} & \textbf{61.2} & \textbf{79.6} & \textbf{76.7} & \textbf{70.9} & \textbf{68.9} \\ 
\bottomrule

\end{tabular}}
\caption{Experimental results of extracting all events on ACE2005-Nest, Genia11, and Genia13, respectively.}
\label{tab:all_res}
\end{table*}

\subsection{Results on Extracting All Events}

Table~\ref{tab:all_res} presents the F1 scores for extracting all events on ACE2005-Nest, Genia11, and Genia13. It can be observed that PerNee outperforms all the baselines. The existence of PEs in the dataset confuses the baselines and affects their performance on all the subtasks. The PerNee model effectively mitigates the impact of dual identities associated with PEs through a two-step process: 1) Firstly, it identifies the candidate PEs within triggers using the trigger recognizer. 2) It subsequently determines the argument identities of the PEs via classifying the relations between trigger pairs with the pivot element recognizer. By disentangling the two identities of PEs through the above two modules, the model achieves enhanced precision in recognizing PEs. Besides, the utilization of label information of event types and argument roles as prompts can help the model get better representations. Thereby, PerNee leads to better performance in most NEE tasks.

In Table~\ref{tab:all_res}, the improvement in AI and AC on ACE2005-Nest is significant while the enhancement on Genia11 and Genia13 is marginal. This may be attributed to the fewer outer-nest event types and a lower proportion of PEs in Genia11 and Genia13. (1) In the Genia11 and Genia13 datasets, there are only 3 and 5 outer-nest event types, with the majority falling into the ``Regulation'' type, while ACE2005-Nest contains 14 outer-nest event types. This implies that the extraction of nested events in Genia11 and Genia13 is relatively easier compared to ACE2005-Nest. (2) the number of PEs accounts for only 19\% and 20\% of all arguments in Genia11 and Genia13, respectively. The AI and AC metrics in Table 3 reflect the extraction performance of all arguments. As our model is primarily optimized for PE recognition, the overall improvement in AI and AC is not significant.

\subsection{Results on Extracting Nested Events}
To further verify that the improvement on ACE2005-Nest, Genia11, and Genia13 are indeed attributed to the enhancement of NEE, we evaluate the performance of PerNee only on the sentences that contain at least one nested event structure in ACE2005-Nest, Genia11, and Genia13. The results are presented in Table~\ref{tab:nee_table}. It can be observed that PerNee significantly surpasses other baselines on all the subtasks on these three datasets, which convincingly verifies its effectiveness. To further analyze the reason, the F1 scores on PEI and PEC are also reported in Table~\ref{tab:nee_table}. It can be noticed that PerNee achieves significant improvement on the PEI and PEC because the designed trigger recognizer and pivot element recognizer can better process the dual identities of PEs.

\begin{table}[ht]
\centering
\resizebox{\linewidth}{!}{
\begin{tabular}{lcccccc}
\toprule
      & TI  & TC  & AI  & AC  & PEI  & PEC \\ \midrule
\multicolumn{5}{l}{$\bullet$ \bf ACE2005-Nest}\\
PLMEE & 66.3 & 64.2 & 46.2 & 45.4 & 28.8 & 28.8 \\
CasEE & 78.3 & 75.7 & 53.5 & 50.9 & 46.1 & 46.1\\
UIE & 78.0 & 75.0 & 54.3 & 51.0 & 46.3 & 46.3 \\
HDGSE & 78.2 & 75.9 & 53.2 & 50.2 & 43.2 & 43.2 \\
OneIE & 74.0 & 71.6 & 49.1 & 48.1 & 24.9 & 24.9 \\
OneEE & 78.5 & 76.9 & 54.4 & 50.2 & 46.8 & 46.8\\
\cdashlinelr{1-7}
PerNee  & \textbf{79.0} & \textbf{77.5} & \textbf{55.7} & \textbf{54.6} & \textbf{49.2} & \textbf{49.2}\\ \midrule \midrule

\multicolumn{5}{l}{$\bullet$ \bf Genia11}\\ 
PLMEE & 70.4 & 68.1 & 65.4 & 64.8 & 54.7 & 53.7 \\
CasEE & 75.0 & 72.2 & 64.4 & 63.8 & 54.0 & 53.2 \\
UIE & 73.5 & 69.5 & 63.5 & 62.9 & 55.1 & 54.4\\
HDGSE & 75.1 & 72.5 & 64.2 & 63.3 & 53.6 & 52.9 \\
OneIE & 73.7 & 69.8 & 64.0 & 63.4 & 53.7 & 52.2\\
OneEE & 75.4 & 73.0 & 65.0 & 63.9 & 57.3 & 55.7\\
\cdashlinelr{1-7}
PerNee  & \textbf{75.7} & \textbf{73.2} & \textbf{66.6} & \textbf{65.9} & \textbf{58.8} & \textbf{56.9}\\ \midrule \midrule

\multicolumn{5}{l}{$\bullet$ \bf Genia13}\\ 
PLMEE & 72.8 & 70.4 & 69.0 & 67.5 & 65.7 & 65.2 \\
CasEE &78.8 & 75.4 & 68.3 & 67.4 & 66.2 & 65.9 \\
UIE & 78.4 & 75.2 & 71.4 & 66.8 & 65.8 & 65.8\\
HDGSE & 79.1 & 76.5 & 70.1 & 67.6 & 65.9 & 65.6 \\
OneIE & 79.0 & 76.0 & 69.2 & 66.1 & 63.9 & 63.5\\
OneEE & 78.6 & \textbf{76.9} & 70.4 & 70.3 & 66.7 & 66.7\\
\cdashlinelr{1-7}
PerNee  & \textbf{79.6} & \textbf{76.9} & \textbf{72.2} & \textbf{70.9} & \textbf{68.1} & \textbf{68.1}\\ \bottomrule

\end{tabular}
}
\caption{Experimental results of extracting nested events on ACE2005-Nest, Genia11, and Genia13, respectively.}
\label{tab:nee_table}
\end{table}

\subsection{Ablation Studies}

To verify the effectiveness of PE recognition for NEE, we treat PEs as regular arguments and recognize them using the same module as the regular argument recognizer rather than the designed PER module in PerNee. The results are denoted as ``w/o PER'' in Table~\ref{tab:ablation}. It shows that removing PER leads to a decrease, particularly in AI, AC, PEI, and PEC metrics. It implies that emphasizing PE recognition and designing a specialized PER module can improve the performance of NEE.

To further verify the effectiveness of identifying PEs via trigger-trigger relation classification, we treat PE recognition as a sequence labeling problem by utilizing a specific FNN with CRF, denoted as ``repl. PER''. It illustrates a significant decline of 27.2\% in both PEI and PEC metrics. This indicates the design of the PER module in PerNee is more effective in recognizing PEs.

Besides, to verify the effectiveness of prompts, we remove them (denoted as ``w/o. pmt''), resulting in performance decreases across all subtasks. It suggests that label information related to event types and argument roles proves beneficial for NEE.

\begin{table}[ht]
\centering
\small
\resizebox{\linewidth}{!}{
\begin{tabular}{lcccccc}
\toprule
 & TI  & TC  & AI  & AC  & PEI  & PEC  \\ \midrule
PerNee & 72.8 & 70.0 & 55.5 & 53.8 & 41.5 & 41.5  \\
\cdashlinelr{1-7}
w/o PER & 72.1 & 69.6 & 54.0 & 52.4 & 35.5 & 35.5 \\
repl. PER & 64.8 & 62.2 & 42.5 & 41.2 & 14.3 & 14.3 \\
w/o pmt. & 71.4 & 69.4 & 53.4 & 52.0 & 40.1 & 40.1\\
\bottomrule

\end{tabular}
}
\caption{Ablation studies on ACE2005-Nest.}
\setlength{\abovecaptionskip}{0pt}
\label{tab:ablation}
\end{table}

\subsection{Detailed Analysis}
To verify the motivation that identifying the dual identities of PEs is crucial for NEE, we compare the F1 scores of PerNee with OneIE\footnote{Here, we adopt OneIE for comparison, as it achieves the second best performance on ACE2005-Nest and allows for the separation of the TI and AI subtasks.} on the following four subtasks, as shown in Figure~\ref{fig:detailed} (a): 1) PE-TI: Identifying the trigger identities of PEs; 2) Identifying regular triggers; 3) AI-PE: Identifying the argument identities of PEs; 4) AI-Reg: Identifying regular arguments. From results shown in Figure~\ref{fig:detailed} (b), we can observe that:

\begin{figure}[h]
    \centering
    \includegraphics[width=1\linewidth]{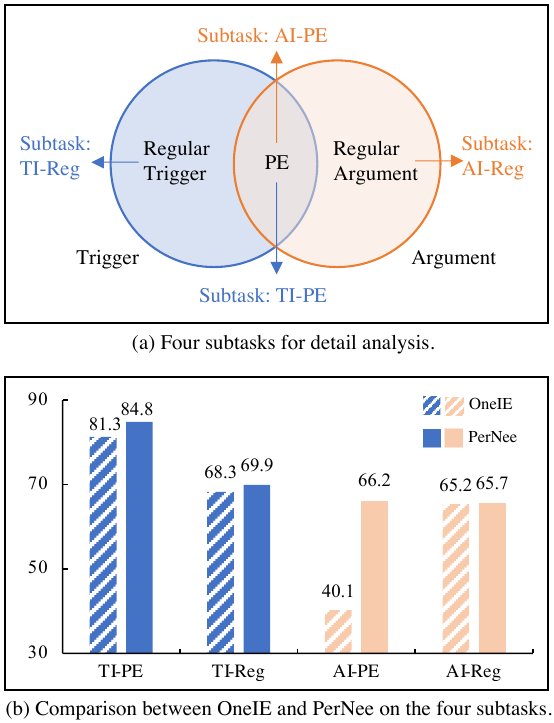}
    \setlength{\abovecaptionskip}{0pt}
    \caption{Detailed analysis on ACE2005-Nest.}
    \setlength{\belowcaptionskip}{0pt}
    \label{fig:detailed}
\end{figure}

1) Identifying the argument identities of PEs is more challenging than the trigger identities of PEs. Both OneIE and PerNee achieve high F1 scores in PE-TI (81.3\% and 84.8\%) but low F1 scores in PE-AI (40.1\% and 66.2\%). It suggests that PEs possess more trigger-like characteristics, making the argument identities of PEs more difficult to be recognized;

2) For OneIE, identifying the argument identities of PEs is more difficult than identifying regular arguments. The F1 score on PE-AI (40.1\%) is significantly lower than that on Reg-AI (65.2\%). It indicates deploying the same module to identify the argument identities of PEs and regular arguments is ineffective;

3) PerNee significantly improves the argument identification of PEs. PerNee outperforms OneIE on all four subtasks. Especially on PE-AI, it increases from 40.1\% to 66.2\%, indicating the effectiveness of the PER module;

4) PerNee exhibits a much smaller performance gap in identifying regular arguments and the argument identities of PEs compared to OneIE. The gap between AI-PE and AI-Reg is 0.5\% for PerNee, while the gap for OneIE is 26.1\%. It indicates that, with the PER module, the AI-PE subtask is not that difficult and can achieve comparable performance with the AI-Reg subtask.

\subsection{Case Study}

\begin{figure}[h]
    \centering
    \includegraphics[width=1\linewidth]{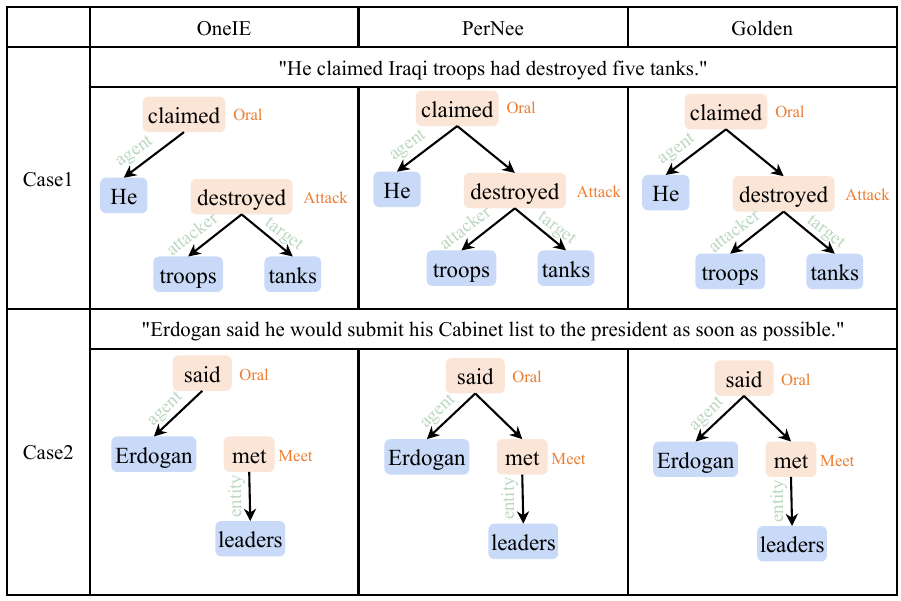}
    \setlength{\abovecaptionskip}{0pt}
    \caption{Case study on the importance of PE recognition for NEE on ACE2005-Nest.}
    \setlength{\belowcaptionskip}{0pt}
    \label{fig:casestudy}
\end{figure}

To demonstrate the importance of Pivot Element (PE) recognition in NEE, two cases are selected from the ACE2005-Nest dataset. As shown in Table~\ref{fig:casestudy}, PerNee correctly extracted the nested events while OneIE, the second-best model in the ACE2005-Nest dataset, failed.

In these cases, PerNee correctly recognized the PEs ("destroyed" in Case 1, "met" in Case 2) and consequently extracted nested events accurately. On the other hand, OneIE successfully recognized most arguments but failed to recognize the argument identities of PEs, resulting in the inability to extract nested events. This highlights the effectiveness of our specifically designed PE recognition module, which enhances the performance of nested event extraction.

\section{Conclusions}

In this paper, we presented PerNee, a new model designed to tackle the Nested Event Extraction (NEE) task by focusing on recognizing Pivot Elements (PEs). PEs are crucial in connecting outer-nest and inner-nest events within a nested structure. Since PEs have dual identities (i.e., trigger identities and argument identities), we employ a trigger recognizer to recognize the trigger identities and utilize a PE recognizer to recognize the argument identities, thus connecting events into nested structures. To enhance NEE performance, the semantic information of event types and argument roles is leveraged through prompt learning. Additionally, we introduced ACE2005-Nest, a new NEE dataset in the generic domain, which systematically categorizes nested events therein. Experimental results demonstrate that the proposed model consistently achieves state-of-the-art performance on ACE2005-Nest, Genia11, and Genia13. Moreover, ablation studies validate the effectiveness of the PE recognizer module and prompts. In the future, we will try to optimize the PE recognizer module and extend our method to handle events with more complex structures, such as multi-level nested events. This will allow us to further advance NEE capabilities and address the challenges posed by intricate event hierarchies.

\vspace{-0.2cm}
\section{Acknowledgments}
The work is supported by the GFKJ Innovation Project, the National Natural Science Foundation of China under grant 62306299, the National Key Research and Development Project of China, the Beijing Academy of Artificial Intelligence under grant BAAI2019ZD0306, the KGJ Project under grant JCKY2022130C039, and the Lenovo-CAS Joint Lab Youth Scientist Project. We appreciate anonymous reviewers for their insightful comments and suggestions.



\section{Bibliographical References}\label{sec:reference}

\bibliographystyle{lrec-coling2024-natbib}
\bibliography{lrec-coling2024-example}


\section{Language Resource References}
\label{lr:ref}
\bibliographystylelanguageresource{lrec-coling2024-natbib}
\bibliographylanguageresource{languageresource}




\end{document}